\documentclass{article}

\usepackage{graphicx}
\usepackage{amsmath}
\usepackage{amssymb}
\usepackage{booktabs}

\usepackage{pifont}
\usepackage{xcolor}
\usepackage{adjustbox}
\usepackage{array}
\usepackage{float}

\usepackage[preprint]{corl_2022} 

\usepackage{hyperref}
\setcitestyle{pagebackref, breaklinks, colorlinks}

\let\oldding\ding 
\renewcommand{\ding}[2][1]{\scalebox{#1}{\oldding{#2}}}

\newcolumntype{P}[1]{>{\centering\arraybackslash}p{#1}}

\title{Pixel-Wise Prediction based Visual Odometry via Uncertainty Estimation}

\author{
  Hao-Wei Chen, Ting-Hsuan Liao, Hsuan-Kung Yang, and Chun-Yi Lee\\
  Elsa Lab, Department of Computer Science\\
  National Tsing Hua University,
  Hsinchu, Taiwan\\
  \texttt{ \{jaroslaw1007, tingforun, hellochick, cylee\}@gapp.nthu.edu.tw} \\
}

\begin{document}\renewcommand{\arraystretch}{1.2}
\maketitle

\begin{abstract}
This paper introduces pixel-wise prediction based visual odometry (PWVO), which is a dense prediction task that evaluates the values of translation and rotation for every pixel in its input observations. PWVO employs uncertainty estimation to identify the noisy regions in the input observations, and adopts a selection mechanism to integrate pixel-wise predictions based on the estimated uncertainty maps to derive the final translation and rotation.  In order to train PWVO in a comprehensive fashion, we further develop a data generation workflow for generating synthetic training data.  The experimental results show that PWVO is able to deliver favorable results.  In addition, our analyses validate the effectiveness of the designs adopted in PWVO, and demonstrate that the uncertainty maps estimated by PWVO is capable of capturing the noises in its input observations.

\end{abstract}


\section{Introduction}
\label{sec::introduction}

Visual odometry (VO) is the process of inferring the correspondence of pixels or features via analyzing the associated camera images, and determining the position and orientation of a robot. Conventionally, the process of VO takes raw RGB images as inputs, derives the correspondence from them, and estimates the changes in translation and rotation of the camera viewpoints between consecutive image frames.  This process enables a robot to derive its entire trajectory from its observed images over a period of time.    In the past decade, there have been a number of VO methods proposed in the literature~\citep{konda2015learning, agrawal2015learning, jayaraman2015learning, deepvo, walch2017image, posecnn, relocnet, laskar2017camera, melekhov2017relative, posenet, posenetwuncertainty, cai2018hybrid, kendall2017geometric, sattler2019understanding, geomapnet, melekhov2017image, naseer2017deep, vlocnet++, valada2017deep, wu2017delving, xue2018guided, d3vo, costante2016exploring, lsvo, flowdometry, wang2019atloc, hamed2020attn, d3vo, ParisottoCZS18, ChenRMLWMT19}. The requirement of estimating such changes from raw RGB images, nevertheless, often causes those methods to suffer from the existence of noises coming from moving objects, as the correspondence extracted from those moving objects might not be directly related to the motion of the camera viewpoint.  This fact leads to degradation in accuracy when performing VO, and limits previous methods to further improve.



In order to deal with the above issue, the objectives of this paper are twofold: (1) validating the fact that the noises from moving objects would degrade the performance of VO, and (2) investigating and proposing an effective approach to mitigate the impacts from them.  Instead of embracing VO frameworks that are based on raw RGB inputs~\citep{konda2015learning, agrawal2015learning, jayaraman2015learning, deepvo, walch2017image, posecnn, relocnet, laskar2017camera, melekhov2017relative, posenet, posenetwuncertainty, cai2018hybrid, kendall2017geometric, sattler2019understanding, geomapnet, melekhov2017image, naseer2017deep, vlocnet++, valada2017deep, wu2017delving, xue2018guided, wang2019atloc, hamed2020attn, d3vo, ParisottoCZS18, ChenRMLWMT19}, in this work, we focus the scope of our discussion on deriving relative translation and rotation from intermediate representations~\citep{costante2016exploring, lsvo, flowdometry, CiarfugliaCVR14, ZhangLKB09}, which include extracted optical flow maps, depth maps, as well as know camera intrinsic. Specifically, we aim to derive and validate our proposed methodology based on perfect intermediate representations, while eliminating the influences of inaccurate features and correspondence from  feature extraction stages.


The first step toward the aforementioned objectives is to develop a mechanism that allows the noises of moving objects from those intermediate representations to be either filtered out or suppressed.  Past researchers have explored three different directions: (i) semantic masks, (ii) attention mechanisms, and (iii) uncertainty estimation. Among them, semantic masks require another separate segmentation model to discover potential moving objects (e.g., cars, pedestrians, etc.)~\citep{DAVO, KanekoIOYA18, BescosFCN18, moav}. Attention mechanisms seek the clues of possible candidates from intermediate representations~\citep{xue2018guided, wang2019atloc, hamed2020attn, ParisottoCZS18, ChenRMLWMT19, DAVO, chen2019selective, feng2020atten, bin2022cross, lee2021atten}. Uncertainty estimation, on the other hand, implicitly captures the noises and enables the models to respond to the measured stochasticity inherent in the observations~\citep{posenetwuncertainty, d3vo, kendall2017what, Klodt2018SupervisingTN, 5509636, 6751290, DAI2021107459, CostanteM20}. All of these three directions have been attempted in the realm of VO.  Nevertheless, the previous endeavors have only been concentrating on predicting a single set of translation and rotation values from their input observations, neglecting the rich information concealed in pixels. In light of these reasons, we propose to employ uncertainty estimation as a means to leverage such information, and further extend VO to be based on pixel-wise predictions. This concept can also be regarded as an implicit form of the attention mechanism.

Different from conventional VO approaches, pixel-wise prediction based VO (or simply "\textit{PWVO}" hereafter) is designed as a dense prediction task, which evaluates the values of translation and rotation for every pixel in its input observations. PWVO first performs predictions for all pixels in the input observations, and then integrates these local predictions into a global one. As PWVO is based on pixel-wise predictions, such a nature allows it to suppress noisy regions  through the usage of uncertainty maps.
The regions with high uncertainty are likely to be noises (e.g., moving objects), and should not be considered in the final global prediction. As a result, a weighting and selection strategy is specifically tailored for PWVO to aggregate the local predictions from its input pixels.



In order to validate the advantages of PWVO, we further develop a data generation workflow, which features a high degree of freedom to generate intermediate representations for PWVO. The workflow is fully configurable, and allows various setups of camera intrinsic, viewpoint motion, extrinsic range, as well as a diverse range of the number, size, and speed for moving objects.  Such a flexibility enables the training data to be comprehensive, and prevents PWVO from overfitting to the setups of a certain existing dataset.
In our experiments, we examine the effectiveness of PWVO in terms of its accuracy, saliency maps, as well as factorized optical maps. We further present a set of ablation analyses to justify the design decisions adopted by PWVO.  The primary contribution of this paper is the introduction of pixel-wise predictions in VO, as well as the dataset generation workflow.

The paper is organized as follows.  Section~\ref{sec:related_work} reviews the related work in the literature.  Section~\ref{sec:methodology} walks through the PWVO framework and its components. Section~\ref{sec::data_generation} describes the dataset generation workflow. Section~\ref{sec:experimental_results} reports the experimental results. Section~\ref{sec:limitation_and_future_direction} discusses the limitations and future directions.  Section~\ref{sec:conclusion} concludes.  The essential background material, hyper-parameter setups, additional results, as well as our reproducible source codes are offered in the supplementary material.






\section{Related Work}
\label{sec:related_work}
Traditional VO methods are based on multiview geometry. According to the algorithms adopted by them, these methods  can be roughly categorized into either feature-based ~\citep{orbslam2, klein2007parallel, viso2} or direct ~\citep{dso, lsdslam, dtam} methods. The former involves correspondences between image frames by using sparse key points, while the latter attempts to recover camera poses by minimizing the image warping photometric errors. Both categories 
suffer from the scale drift issue if the absolute depth scale is unknown.  
The authors in~\citep{agrawal2015learning, jayaraman2015learning} pioneered the usage of convolutional neural networks (CNNs) to learn camera pose estimation with various types of loss functions and model architectures.  The authors in~\citep{walch2017image, posenet, geomapnet, melekhov2017image} proposed to directly regress six degrees of freedom (6-DoF) camera poses through weighted combinations of position and orientation errors. Methods employing geometric reprojection errors~\citep{kendall2017geometric} and visual odometry constraints~\citep{posenet, vlocnet++, valada2017deep} were also introduced in the literature.  Moreover, the authors in~\citep{naseer2017deep, wu2017delving} attempted to train their models with extra synthetic data, while the authors in~\citep{deepvo} trained their VO models from videos using recurrent neural networks.  Furthermore, the technique proposed in~\citep{walch2017image} adopted an end-to-end approach that merges camera inputs with inertial measurement unit (IMU) readings using an LSTM network. Most of these image-based localization methods perform VO via retriving information from input images.  As a result, feature extraction is critical to their performance, which in turn impacts their generalizability to challenging scenarios.
To reduce the impacts from feature extraction stages, some researchers~\cite{costante2016exploring, lsvo, flowdometry} proposesed to use optical flow maps as inputs instead of RGB images. The authors in~\cite{lsvo} further proposed to utilize autoencoder networks to learn a better representations for their optical flow maps. To eliminate the noises from input observations, a number of researchers have also investigated attention based VO methods~\citep{xue2018guided, wang2019atloc, hamed2020attn, ParisottoCZS18, DAVO, chen2019selective, feng2020atten, bin2022cross} and uncertainty based VO methods~\citep{d3vo, Klodt2018SupervisingTN, DAI2021107459, CostanteM20} to mitigate the impacts of moving objects from their input observations.





\section{Methodology}
\label{sec:methodology}

In this section, we first formally define our problem formation, and provide an overview of the PWVO framework. Next, we describe the two constituent stages in PWVO. Finally, we introduce the refinement strategy as well as the total loss term, and elaborate on the rationale behind them.

\subsection{Problem Formation}
\label{subsec::problem_formulation}

Given an optical flow field $\mathbf{F}^{total}_{i} \in \mathbb{R}^{H \times W \times 2}$, depth maps $(\mathbf{D}_{i}, \mathbf{D}_{i+1}) \in \mathbb{R}^{H \times W \times 1}$, two dimensional pixel coordinates $\mathbf{x} \in \mathbb{R}^{H \times W \times 2}$, and the camera intrinsic $\mathbf{K}_{i} \in \mathbb{R}^{3 \times 3}$, the proposed PWVO framework aims at predicting a tuple of camera rotation $\tilde{\gamma}_{i} \in  \mathbb{R}^{3}$ and translation $\tilde{\varphi}_{i} \in \mathbb{R}^{3}$, where $i$ denotes the frame index, and $H, W$ represent the height and width of the input frames, respectively.  

To achieve the above objective, PWVO first performs pixel-wise predictions of the camera motion $(\tilde{\mathcal{R}}_{i}, \tilde{\mathcal{T}}_{i})$, where $\tilde{\mathcal{R}}_{i} \in \mathbb{R}^{H \times W \times 3}$ and $\tilde{\mathcal{T}}_{i} \in \mathbb{R}^{H \times W \times 3}$ correspond to the pixel-wise rotation and translation maps, respectively.  In addition, PWVO further generates an uncertainty map tuple $(\tilde{\mathcal{U}}^{\mathcal{R}}_{i}, \tilde{\mathcal{U}}^{\mathcal{T}}_{i})$, where $\tilde{\mathcal{U}}^{\mathcal{R}}_{i} \in \mathbb{R}^{H \times W \times 3}$ and $\tilde{\mathcal{U}}^{\mathcal{T}}_{i} \in \mathbb{R}^{H \times W \times 3}$, to reflect the uncertainty (i.e., noises) in the input observations. The predicted $(\tilde{\mathcal{R}}_{i}, \tilde{\mathcal{T}}_{i})$ are then used along with  ($\tilde{\mathcal{U}}^{\mathcal{R}}_{i}, \tilde{\mathcal{U}}^{\mathcal{T}}_{i}$) to derive the final ($\tilde{\gamma}_{i}, \tilde{\varphi}_{i}$).

\subsection{Overview of the PWVO Framework}
\label{subsec:framework_overview}

\begin{figure}[t]
  \centering
  \includegraphics[width=.95\textwidth]{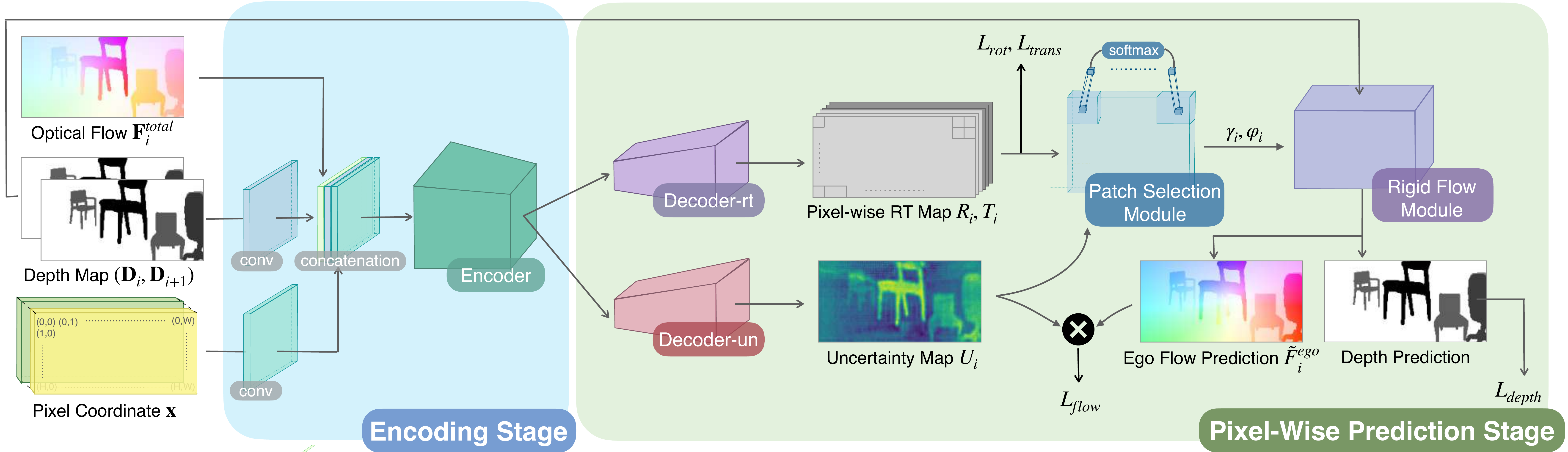}
  \caption{An overview of the proposed PWVO framework.}
  \label{fig:PWVO_framework_overview}
\end{figure}

Fig.~\ref{fig:PWVO_framework_overview} illustrates the proposed PWVO framework, which consists of two stages: (i) an encoding stage and (ii) a pixel-wise prediction stage. 
The function of the encoding stage is to encode the inputs (i.e., $\mathbf{F}^{total}_{i}, \mathbf{D}_{i}, \mathbf{D}_{i+1}$, and $\mathbf{x}$) by a series of convolutional operations into a feature embedding $\tilde{\psi}$, which bears the motion information concealed in the inputs.  This embedding is then forwarded to the pixel-wise prediction stage to generate $(\tilde{\mathcal{R}}_{i}, \tilde{\mathcal{T}}_{i})$ and $\tilde{\mathcal{U}}_{i}$ by two separated branches. The uncertainty map $\tilde{\mathcal{U}}_{i}$ is utilized to reflect the noises contained in the inputs.  On the other hand, the predicted $(\tilde{\mathcal{R}}_{i}, \tilde{\mathcal{T}}_{i})$ and $\tilde{\mathcal{U}}_{i}$ are later fed into a selection module, which employs a patch selection procedure to mitigate the impacts of the regions that may potentially contain moving objects by referring to $\tilde{\mathcal{U}}_{i}$, and aggregates the weighted predictions to derive the final $(\tilde{\gamma}_{i}, \tilde{\varphi}_{i})$.  
In order to further refine the predicted $(\tilde{\gamma}_{i}, \tilde{\varphi}_{i})$, PWVO additionally reconstructs an ego flow prediction $\tilde{\mathbf{F}}^{ego}_{i} \in \mathbb{R}^{H \times W \times 2}$ as well as a depth map $\tilde{\mathbf{D}}_{i+1} \in \mathbb{R}^{H \times W \times 1}$ based on $(\tilde{\gamma}_{i}, \tilde{\varphi}_{i})$.  The reconstructed $\tilde{\mathbf{F}}^{ego}_{i}$, $\tilde{\mathbf{D}}_{i+1}$, and $(\tilde{\gamma}_{i}, \tilde{\varphi}_{i})$ are then compared against their corresponding ground truth labels $\mathbf{F}^{ego}_{i}$, $\mathbf{D}_{i+1}$, and $(\gamma_{i}, \varphi_{i})$ respectively to optimize the model parameters $\theta$ in PWVO through backward propagation.






\subsection{Encoding Stage}
\label{subsec:stage_I}
The encoding stage first forwards $\mathbf{F}^{total}_{i}, \mathbf{D}_{i}, \mathbf{D}_{i+1}$, and $\mathbf{x}$ through three distinct branches, which are then followed by an encoder to transform the concatenated feature embeddings from the outputs of the three branches into $\tilde{\psi}$. In addition to the flow and depth information, the last branch of the encoding stage is designed to encapsulate the positional clues of $\mathbf{x}$.  This encapsulation process allows PWVO to learn translation dependence~\citep{coordconv}, and is expressed as the following equation:
\begin{equation}
\label{eq::coordinate_transform}
    \mathbf{X}^{p}_{i} = \mathbf{K}_{i}^{-1}\mathbf{x}^{p}_{i}, \qquad \mathbf{K}_{i} \in \mathbb{R}^{3 \times 3}, \mathbf{x}^{p}_{i} \in \mathbb{R}^{3 \times 1}, \mathbf{N} \in \mathbb{R}^{(H \times W)},
\end{equation}
where $p$ denotes the pixel index, $\mathbf{X}^{p}_{i}$ represents the three dimensional coordinates in the film space, and $\mathbf{N}$ is the pixel number $H \times W$. Eq.~(\ref{eq::coordinate_transform}) reveals that the third branch of the encoding stage allows the information of the camera intrinsic and $\mathbf{x}$ to be carried over to $\tilde{\psi}$. This can potentially benefit PWVO to possess better understanding about positioning and translation dependence, as the motion features provided by $\mathbf{F}^{total}_{i}$ and the coordinate information offered by $\mathbf{x}$ can complement each other.

\subsection{Pixel-Wise Prediction Stage} 

The pixel-wise prediction stage first utilizes two decoders \texttt{Decoder-rt} and \texttt{Decoder-un}, as depicted in Fig.~\ref{fig:PWVO_framework_overview}, to upsample $\tilde{\psi}$ to $(\tilde{\mathcal{R}}_{i}, \tilde{\mathcal{T}}_{i})$ and $(\tilde{\mathcal{U}}^{\mathcal{R}}_{i}, \tilde{\mathcal{U}}^{\mathcal{T}}_{i})$, respectively. In the following subsections, we elaborate on the details of the distribution learning procedure as well as the selection module.

\subsubsection{Distribution Learning}
\label{subsubsec:uncertainty}

The distribution learning procedure in PWVO aims at learning the posterior probability distributions of rotation and translation at each pixel. Assume that the noises are modeled as Laplacian, this procedure can be carried out by leveraging the concept of heteroscedastic aleatoric uncertainty of deep neural networks (DNNs) discussed in~\citep{kendall2017what}. More specifically, $(\tilde{\mathcal{R}}_{i}, \tilde{\mathcal{T}}_{i})$ and $(\tilde{\mathcal{U}}^{\mathcal{R}}_{i}, \tilde{\mathcal{U}}^{\mathcal{T}}_{i})$ are learned together by minimizing the loss terms $\mathcal{L}^{R}_{i}$ and $\mathcal{L}^{T}_{i}$ for all pixels, which can be expressed as:
\begin{align}
  &\begin{aligned}
    \label{eq::loss_R}
    \hat{\mathcal{L}}^{\mathcal{R}}_{i} = \frac{1}{\mathbf{N}} \sum^{\mathbf{N}}_{p=1} \frac{\mathcal{E}^{\mathcal{R}}(\gamma_{i}, \tilde{\mathcal{R}}^{p}_{i})}{\tilde{\mathcal{U}}^{\mathcal{R}}_{i}(p)}   + \log (\tilde{\mathcal{U}}^{\mathcal{R}}_{i}(p)), \quad \mathcal{E}^{\mathcal{R}}(x, y) = \Vert x - y \Vert,
  \end{aligned}\\
  &\begin{aligned}
    \label{eq::loss_T}
    \hat{\mathcal{L}}^{\mathcal{T}}_{i} = \frac{1}{\mathbf{N}} \sum^{\mathbf{N}}_{p=1} \frac{\mathcal{E}^{\mathcal{T}}(\varphi_{i}, \tilde{\mathcal{T}}^{p}_{i})}{\tilde{\mathcal{U}}^{\mathcal{T}}_{i}(p)}   + \log (\tilde{\mathcal{U}}^{\mathcal{T}}_{i}(p)), \quad \mathcal{E}^{\mathcal{T}}(x, y) = \Vert \langle x \rangle - \langle y \rangle \Vert + ({\Vert x \Vert}_{2} - {\Vert y \Vert}_{2})^{2},
  \end{aligned}
\end{align}
where $\langle \cdot \rangle$ denotes the Euclidean normalization vector, and ($\tilde{\mathcal{R}}^{p}_{i}, \tilde{\mathcal{T}}^{p}_{i}$) and $(\tilde{\mathcal{U}}^{\mathcal{R}}_{i}(p), \tilde{\mathcal{U}}^{\mathcal{T}}_{i}(p))$ represent the means and variances of the probability distributions of rotation and translation at pixel $p$, respectively. Please note that $(\tilde{\mathcal{U}}^{\mathcal{R}}_{i}(p), \tilde{\mathcal{U}}^{\mathcal{T}}_{i}(p))$ are learned implicitly, and the second terms in Eqs.~(\ref{eq::loss_R}) and~(\ref{eq::loss_T}) regulate the scales of them. The loss functions allow PWVO to adapt its uncertainty estimation for different pixels, which in turn enhance its robustness to noisy data or erroneous labels.

In practice, \texttt{Decoder-un} is modified to predict log variance, and  Eqs.~(\ref{eq::loss_R})-(\ref{eq::loss_T}) are reformulated as:
\begin{align}
  &\begin{aligned}
    \hat{\mathcal{L}}^{\mathcal{R}}_{i} = \frac{1}{\mathbf{N}}\sum^{\mathbf{N}}_{p=1} \exp{(-\tilde{s}^{\mathcal{R}}_{i}(p))} \cdot \mathcal{E}^{\mathcal{R}} (\gamma_{i}, \tilde{\mathcal{R}}^{p}_{i}) + \tilde{s}^{\mathcal{R}}_{i}(p), \quad \tilde{s}^{\mathcal{R}}_{i}(p) = \log (\tilde{\mathcal{U}}^{\mathcal{R}}_{i}(p)).
  \end{aligned}\\
  &\begin{aligned}
    \hat{\mathcal{L}}^{\mathcal{T}}_{i} = \frac{1}{\mathbf{N}}\sum^{\mathbf{N}}_{p=1} \exp{(-\tilde{s}^{\mathcal{T}}_{i}(p))} \cdot \mathcal{E}^{\mathcal{T}} (\varphi_{i}, \tilde{\mathcal{T}}^{p}_{i}) + \tilde{s}^{\mathcal{T}}_{i}(p), \quad \tilde{s}^{\mathcal{T}}_{i}(p) = \log (\tilde{\mathcal{U}}^{\mathcal{T}}_{i}(p)).
  \end{aligned}
\end{align}
This modification allows the training progress of PWVO to be stabler than the original formulation, as it avoids errors resulted from division by zero. Moreover, the exponential mapping also enables PWVO to regress unconstrained $\tilde{s}^{\mathcal{R}}_{i}(p)$ and $\tilde{s}^{\mathcal{T}}_{i}(p)$, as exp$(\cdot)$ guarantees the outputs to be positive.



\subsubsection{Selection Module}
\label{subsubsec:selection_module}

The function of the selection module is to derive $(\tilde{\gamma}_{i}, \tilde{\varphi}_{i})$ from ($\tilde{\mathcal{R}}_{i}, \tilde{\mathcal{T}}_{i}$) and $(\tilde{\mathcal{U}}^{\mathcal{R}}_{i}(p), \tilde{\mathcal{U}}^{\mathcal{T}}_{i}(p))$.  It adopts a hierarchical derivation procedure, in which
the $H \times W$ pixels of a frame are first grouped into $h \times w$ patches of size $k \times k$ pixels, where  $h = H / k, w = W / k$. The rotation, translation, and uncertainty maps for each patch can then be directed extracted from the original ones, and are represented as $(\tilde{\mathfrak{r}}_{l, m}, \tilde{\mathfrak{t}}_{l, m})$ and $(\tilde{\mathfrak{u}}^{\mathcal{R}}_{l, m}, \tilde{\mathfrak{u}}^{\mathcal{T}}_{l, m})$, where  $l$ and $m$ denote the row and the column indices of a certain patch. The selection module next selects the pixel with the lowest uncertainty value within each patch, represented as:  $p^{\mathcal{R}}_{l, m} = argmin(\tilde{\mathfrak{u}}^{\mathcal{R}}_{l, m}), p^{\mathcal{T}}_{l, m} = argmin(\tilde{\mathfrak{u}}^{\mathcal{T}}_{l, m})$, where $p^{\mathcal{R}}_{l, m}$ and $p^{\mathcal{T}}_{l, m}$ are the pixel indices corresponding to patch $(l, m)$.  They are used to derive the final global $(\tilde{\gamma}_{i}, \tilde{\varphi}_{i})$ as:
\begin{align}
  & \begin{aligned}
    \tilde{\gamma}_{i} = \sum^{h}_{l=1} \sum^{w}_{m=1} \mathcal{W}^{\mathcal{R}}_{l, m} \cdot \tilde{\mathfrak{r}}_{l, m}(p^{\mathcal{R}}_{l, m}), \quad \mathcal{W}^{\mathcal{R}}_{l, m} = \frac{\exp (\tilde{\mathfrak{u}}^{\mathcal{R}}_{l, m}(p^{\mathcal{R}}_{l, m}))}{\sum^{h}_{l = 1}\sum^{w}_{m = 1} \exp (\tilde{\mathfrak{u}}^{\mathcal{R}}_{l, m}(p^{\mathcal{R}}_{l, m}))}.
  \end{aligned}\\
  & \begin{aligned}
    \tilde{\varphi}_{i} = \sum^{h}_{l=1} \sum^{w}_{m=1} \mathcal{W}^{\mathcal{T}}_{l, m} \cdot \tilde{\mathfrak{t}}_{l, m}(p^{\mathcal{T}}_{l, m}), \quad \mathcal{W}^{\mathcal{T}}_{l, m} = \frac{\exp (\tilde{\mathfrak{u}}^{\mathcal{T}}_{l, m}(p^{\mathcal{T}}_{l, m}))}{\sum^{h}_{l = 1}\sum^{w}_{m = 1} \exp (\tilde{\mathfrak{u}}^{\mathcal{T}}_{l, m}(p^{\mathcal{T}}_{l, m}))}.
  \end{aligned}
\end{align}

The main advantage of the above hierarchical procedure is that it enforces $(\tilde{\gamma}_{i}, \tilde{\varphi}_{i})$ to be derived from all the patches from the entire image instead of only concentrating on a certain local region.

\subsection{Refinement and Total Loss Adopted by PWVO}
\label{subsec:loss_function}
In order to further refine the predicted $(\tilde{\gamma}_{i}, \tilde{\varphi}_{i})$, PWVO additionally reconstructs $\tilde{\mathbf{F}}^{ego}_{i}$ and $\tilde{\mathbf{D}}_{i+1}$ based on $(\tilde{\gamma}_{i}, \tilde{\varphi}_{i}$), and compare them against their ground truth labels  $\mathbf{F}^{ego}_{i}$ and $\mathbf{D}_{i+1}$ to optimize the model parameters in both stages of PWVO. The total loss of PWVO can thus be formulated as:
\begin{equation}\label{eq::total_loss}
    \mathcal{L}_{i}^{Total} = \hat{\mathcal{L}^{\mathcal{R}}_{i}} + \hat{\mathcal{L}^{\mathcal{T}}_{i}} + \hat{\mathcal{L}^{\mathcal{D}}_{i}} + 
    \hat{\mathcal{L}^{\mathcal{F}}_{i}},
\end{equation}
where $\hat{\mathcal{L}^{\mathcal{D}}_{i}}$ and $\hat{\mathcal{L}^{\mathcal{F}}_{i}}$ represent the loss functions for ego flow and depth reconstruction, respectively. The detailed formulation of the loss functions $\hat{\mathcal{L}^{\mathcal{D}}_{i}}$ and  $\hat{\mathcal{L}^{\mathcal{F}}_{i}}$ are offered in the supplementary material.

The rationale behind the additional two loss terms (i.e., $\hat{\mathcal{L}^{\mathcal{D}}_{i}}$ and $\hat{\mathcal{L}^{\mathcal{F}}_{i}}$) in Eq.~(\ref{eq::total_loss}) can be explained from two different perspectives. First, re-projection from 2D coordinates to 3D coordinates might potentially cause ambiguity issues if depth information is not taken into consideration. Second, due to the pixel-wise design of PWVO, the optimization target should be different for each pixel coordinate, as the position and depth is different for each pixel. As a result, only optimizing $(\tilde{\mathcal{R}}_{i}, \tilde{\mathcal{T}}_{i})$ without considering the depth and the positional information could be insufficient. These are the reasons that $\hat{\mathcal{L}^{\mathcal{D}}_{i}}$ and $\hat{\mathcal{L}^{\mathcal{F}}_{i}}$ are included in the final optimization target of PWVO. In Section~\ref{subsub:ego_flow_reconstruct}, an ablation analysis is provided to validate the effectiveness of these two additional loss terms.

Interestingly, part of the design of the optimization target $\hat{\mathcal{L}^{\mathcal{F}}_{i}}$ is similar to that of the well-known perspective-n-point (PnP)~\citep{pnp} approach, and can be viewed as a variant of it. This is because the re-projection error $\mathbf{E}_{i}$ in $\hat{\mathcal{L}^{\mathcal{F}}_{i}}$ can be formulated as the L2 loss between $\tilde{\mathbf{F}}^{ego}_{i}$ and $\mathbf{F}^{ego}_{i}$, expressed as:

\begin{equation}
    \mathbf{E}_{i} = \sum^{\mathbf{N}}_{p=1} \Vert \tilde{\mathbf{F}}^{ego}_{i}(p) - \mathbf{F}^{ego}_{i}(p) \Vert_{2} =
    \sum^{\mathbf{N}}_{p=1} \Vert (\frac{1}{\mathbf{D}_{i+1}} \mathbf{K}_{i} \mathbf{M}_{i} \mathbf{X}^{p}_{i} - \mathbf{x}^{p}) - (\mathbf{F}^{ego}_{i}(p)) \Vert_{2}.
\end{equation}
As a result, $\hat{\mathcal{L}^{\mathcal{F}}_{i}}$ implicitly introduces geometric constraints for improving the estimation of $(\tilde{\gamma}_{i}, \tilde{\varphi}_{i})$.




\section{Data Generation Workflow}
\label{sec::data_generation}
\begin{figure}[t]
  \centering
  \includegraphics[width=.95\textwidth]{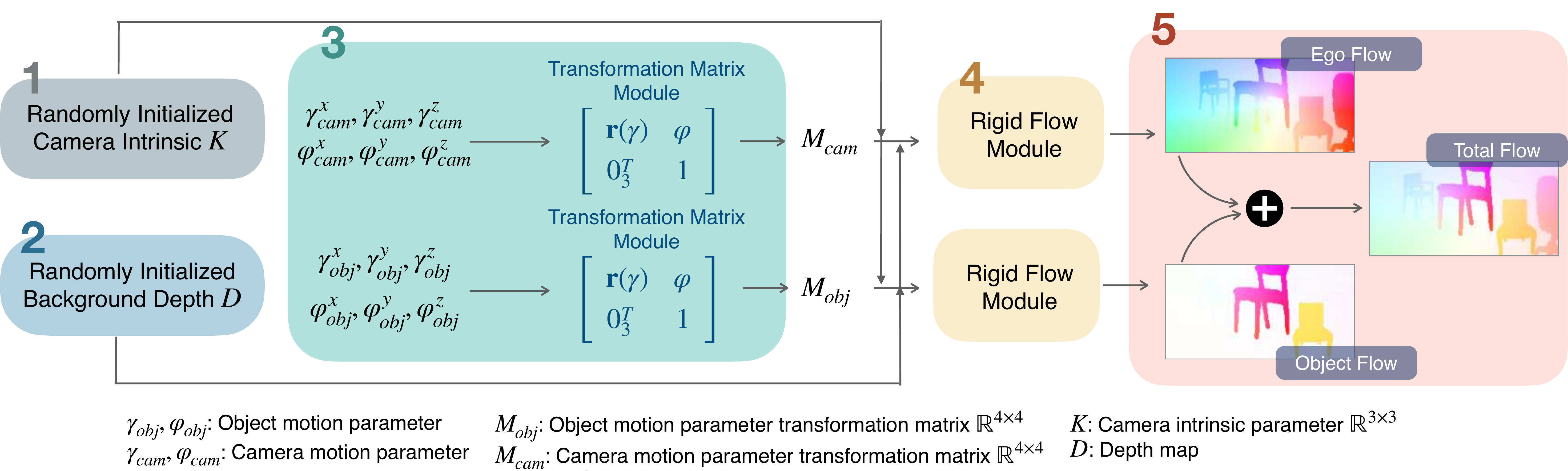}
  \caption{An illustration of the data generation workflow. }
  \label{fig:dataset}
 \end{figure}

In this section, we introduce our data generation workflow for generating synthetic training data.  The workflow is developed to be fully configurable,  with an aim to provide various setups of camera intrinsic $\mathbf{K}$, background depth $\mathbf{D}_{t}$, where $t$ represents the current timestep, as well as diverse combinations of the motions of the camera and the moving objects.  Fig.~\ref{fig:dataset} illustrates the data generation workflow, which consists of five distinct steps.   \textbf{Step~1} initializes a $\mathbf{K}$ by sampling from a distribution, which is detailed in the supplementary material. \textbf{Step~2} randomly generates a $\mathbf{D}$ based on the focal lengths in $\mathbf{K}$.  \textbf{Step~3} randomly initializes the rotation $\gamma$ and translation $\varphi$ for the camera and a set of moving objects, and use them to derive their corresponding transformation matrices $\mathbf{M}$. Subsequently, in \textbf{Step~4}, each transformation matrix is forwarded along with $\mathbf{K}$ and $\mathbf{D}_{t}$ to a rigid flow module to derive a rigid flow map $\mathbf{{F}}^{rigid}$.  The derivation procedure can be formulated as: 
 \begin{equation}\label{eq1}
     \mathbf{F}^{rigid} = \frac{1}{\mathbf{D}_{t+1}}\mathbf{K}\mathbf{M}(\mathbf{D}_{t})\mathbf{K}^{-1}\mathbf{x} - \mathbf{x}, \quad \mathbf{M} = \begin{bmatrix}
        \mathbf{r}(\gamma) & \varphi\\
        0^{\intercal}_{3} & 1
     \end{bmatrix},
 \end{equation}
where $\mathbf{r}(\cdot)$ represents the function that transforms a Euler angle to a rotation matrix. Please note that the rigid flow map derived from the camera motion is referred to as the ego flow map $\mathbf{{F}}^{ego}$, while the rigid flow maps derived from the motions of the moving objects correspond to the object flow maps $\mathbf{{F}}^{obj}$. Finally, \textbf{Step~5} combines all the flow maps together to obtain the total flow map $\mathbf{{F}}^{total}$.  The generated $\mathbf{{F}}$, $\mathbf{K}$, $\mathbf{D}$, and $\mathbf{M}$ are all used in Eq.~(\ref{eq::total_loss}) for training the model parameters in PWVO.



\section{Experimental Results}
\label{sec:experimental_results}
In this section, we present the setups, the quantitative and qualitative results, and a set of analyses.

\subsection{Experimental Setup}
In order to evaluate the performance of PWVO, the effectness of each component of PWVO, as well as the proposed data generation workflow, we design a number of experiments based on the following experimental setups. We train PWVO on a dataset of 100k samples generated by the proposed data generation workflow, and evaluate the trained PWVO on the validation sets of Sintel~\citep{sintel}. The detailed configuration for generating the training dataset is provided in the supplementary material. The trained PWVO is then evaluated using the following metrics: (1) the average rotation error \textit{$R_{err}$} and translation error \textit{$T_{err}$}, which are defined as the L1 error between $(\gamma_{i}, \varphi_{i})$ and $(\tilde{\gamma}_{i}, \tilde{\varphi}_{i})$; (2) the end-point-error (EPE) for measuring the quality of the reconstructed $\tilde{\mathbf{F}}^{ego}_{i}$, which can serve as another metric for evaluating the performance of VO. In our experiments, EPE is defined as the average L1 error between $\mathbf{F}^{ego}_{i}$ and $\tilde{\mathbf{F}}^{ego}_{i}$, which is commonly adopted by flow estimation methods.



\begin{table}[t]
    \centering
    \footnotesize	
    \begin{minipage}[t]{.49\textwidth}
        \centering
        \resizebox{\textwidth}{!}{%
        \begin{tabular}[t]{l|c|c|c}
            \toprule
             & EPE & $R_{err}$ & $T_{err}$\\
            \hline
            VONet~\citep{tartan} & 0.909 & 0.110 & 0.061 \\
            VONet + \textit{self-att.}~\citep{hamed2020attn} & 0.894 & 0.117 & 0.076 \\
            PWVO (\textit{naive}) & 0.829 & 0.091 & 0.061 \\
            PWVO & \textbf{0.626} & \textbf{0.081}& \textbf{0.043} \\
            \toprule
        \end{tabular}}
        \caption{Comparison of PWVO and the baselines in terms of \textit{$R_{err}$}, \textit{$T_{err}$}, and EPE.  It can be observed that PWVO outperforms the two baselines as well as PWVO (\textit{naive}) by noticeable margins.
        }
        \label{table:main_result}
    \end{minipage}\hfill
    \begin{minipage}[t]{.49\textwidth}
        \centering
        \resizebox{\textwidth}{!}{%
        \begin{tabular}[t]{l|c|c|c}
            \toprule
             Configurations & EPE & $R_{err}$ & $T_{err}$\\ \hline
             VONet~\citep{tartan} & 0.909 & 0.110 & 0.061 \\ 
             + pixel-wise & 1.07 & 0.106 & 0.055\\
             + $\mathcal{U}^{\mathcal{R}}$ + $\mathcal{U}^{\mathcal{T}}$(i.e., PWVO (\textit{naive})) & 0.829 & 0.091 & 0.061\\
             + $\mathcal{U}^{\mathcal{D}}$ + $\mathcal{U}^{\mathcal{F}}$ & 0.766 & 0.087 & 0.062\\
             + Selection Module (i.e., PWVO) & \textbf{0.626} & \textbf{0.081} & \textbf{0.043}\\
            \toprule
        \end{tabular}}
        \caption{Ablation study for the effectiveness of the components in PWVO. The pixel-wise predictions are averaged to generate final outputs by default, if the selection module is not adopted.}
        \label{table:ablative_study}
    \end{minipage}
\end{table}


\subsection{Quantitative Results}
In this section, we first compare PWVO against two baselines that adopt different mechanisms for suppressing noisy regions in the input observations. Next, we ablatively examine the effectiveness of the components in PWVO. Finally, we validate the importance of $\hat{\mathcal{L}^{\mathcal{F}}_{i}}$ in optimizing PWVO.

\subsubsection{Comparison of PWVO and the Baselines}

In this experiment, we compare PWVO against VONet~\citep{tartan} and its variant with self-attention mechanism~\citep{hamed2020attn}, which are employed as the baselines and are denoted as \textit{VONet} and \textit{VONet+self-att.}, respectively. VONet is implemented using a similar architecture as the encoding stage of PWVO, and directly predicts rotation and translation from $\tilde{\psi}$. For PWVO, we consider two different configurations: PWVO~(\textit{naive}) and PWVO, where the former directly derives $(\tilde{\gamma}_{i}, \tilde{\varphi}_{i})$ from ($\tilde{\mathcal{R}}_{i}, \tilde{\mathcal{T}}_{i}$) by performing an average operation instead of using the selection module. Moreover, PWVO~(\textit{naive}) does not take into account the uncertainty maps in its $\hat{\mathcal{L}^{\mathcal{D}}_{i}}$ and $\hat{\mathcal{L}^{\mathcal{F}}_{i}}$, and simply resorts to L1 and L2 losses when optimizing its $\tilde{\mathbf{D}}_{i+1}$ and $\tilde{\mathbf{F}}^{ego}_{i}$, respectively.  The comparison results are shown in Table~\ref{table:main_result}. It can be observed that both versions of PWVO are able to outperform the baselines in terms of \textit{$R_{err}$}, \textit{$T_{err}$}, and EPE, validating the effectiveness of the pixel-wise prediction mechanism.





\subsubsection{Ablation Study for the Effectiveness of the Components in PWVO}
\label{subsub:ablation_study}
In this section, we ablatively examine the effectiveness of each component of PWVO by gradually incorporating them into the framework.   The results are reported in Table~\ref{table:ablative_study}.  Please note that $\mathcal{U}^{\mathcal{D}}$ and $\mathcal{U}^{\mathcal{F}}$ are the uncertainty maps used in $\hat{\mathcal{L}}^{\mathcal{D}}$ and $\hat{\mathcal{L}}^{\mathcal{F}}$, which are detailed in the supplementary material. It can be observed that, when simply incorporating the pixel-wise design into VONet without uncertainty estimation, the performance degrades slightly. However, when incorporating both the pixel-wise design and the uncertainty estimation for $\mathcal{U}^{\mathcal{R}}$ and $\mathcal{U}^{\mathcal{T}}$, PWVO~(\textit{naive}) becomes capable of outperforming VONet, indicating that the proposed pixel-wise design is complementary to the uncertainty estimation strategy. The results also reveal that the performance of the model keeps increasing when each new component is added, validating that all them are crucial for PWVO.

\subsubsection{Importance of the Additional Reconstruction Loss $\hat{\mathcal{L}}^{\mathcal{F}}$}

\label{subsub:ego_flow_reconstruct}

In this section, we validate the importance of the reconstruction loss discussed in Section~\ref{subsec:loss_function}. Our hypothesis is that incorporating an additional reconstruction loss term $\hat{\mathcal{L}}^{\mathcal{F}}$ could introduce geometric constraints for improving the performance of PWVO. To validate the assumption, we train the baselines and PWVO with and without the reconstruction loss $\hat{\mathcal{L}}^{\mathcal{F}}$ and analyze their results, which are summarized in Table~\ref{table:ego_reconstruction}. It can be observed that, with the help of the reconstruction loss, nearly all the approaches are able to further enhance their performance in terms of EPE, $\textit{R}_{err}$, and $\textit{T}_{err}$. This evidence thus supports our hypothesis that optimizing PWVO with $\hat{\mathcal{L}}^{\mathcal{F}}$ is indeed beneficial.






\newcommand\Tstrut{\rule{0pt}{2.6ex}}   

\begin{table}[t]
\centering
\scriptsize
\begin{tabular}{l|cc|cc|cc}
\toprule
                  & \multicolumn{2}{c|}{EPE}                      & \multicolumn{2}{c|}{$\textit{R}_{err}$}                  & \multicolumn{2}{c}{$\textit{T}_{err}$}                         \\ \hline
                  & \multicolumn{1}{c|}{\Tstrut w/ $\hat{\mathcal{L}}^{\mathcal{F}}$}         & w/o $\hat{\mathcal{L}}^{\mathcal{F}}$ & \multicolumn{1}{c|}{w/ $\hat{\mathcal{L}}^{\mathcal{F}}$}        & w/o $\hat{\mathcal{L}}^{\mathcal{F}}$ & \multicolumn{1}{c|}{w/ $\hat{\mathcal{L}}^{\mathcal{F}}$}        & w/o $\hat{\mathcal{L}}^{\mathcal{F}}$       \\ \hline
VONet~\citep{tartan} & \multicolumn{1}{c|}{0.909}          & 1.276   & \multicolumn{1}{c|}{0.110}          & 0.113    & \multicolumn{1}{c|}{\textbf{0.061}} & 0.069 \\
VONet + \textit{self-att.}~\citep{hamed2020attn} & \multicolumn{1}{c|}{0.894}          & 1.312       & \multicolumn{1}{c|}{0.117}         &    0.118 & \multicolumn{1}{c|}{0.076}         &     0.065         \\
PWVO              & \multicolumn{1}{c|}{\textbf{0.829}} & 1.241   & \multicolumn{1}{c|}{\textbf{0.091}} & 0.144    & \multicolumn{1}{c|}{\textbf{0.061}} & 0.095          \\ 
\toprule
\end{tabular}
\caption{Validation of the effectiveness of the additional loss term $\hat{\mathcal{L}}^{\mathcal{F}}$.}
\label{table:ego_reconstruction}
\end{table}

\begin{figure}[t]
    \begin{minipage}[t]{.49\textwidth}
        \centering
        \includegraphics[width=\textwidth]{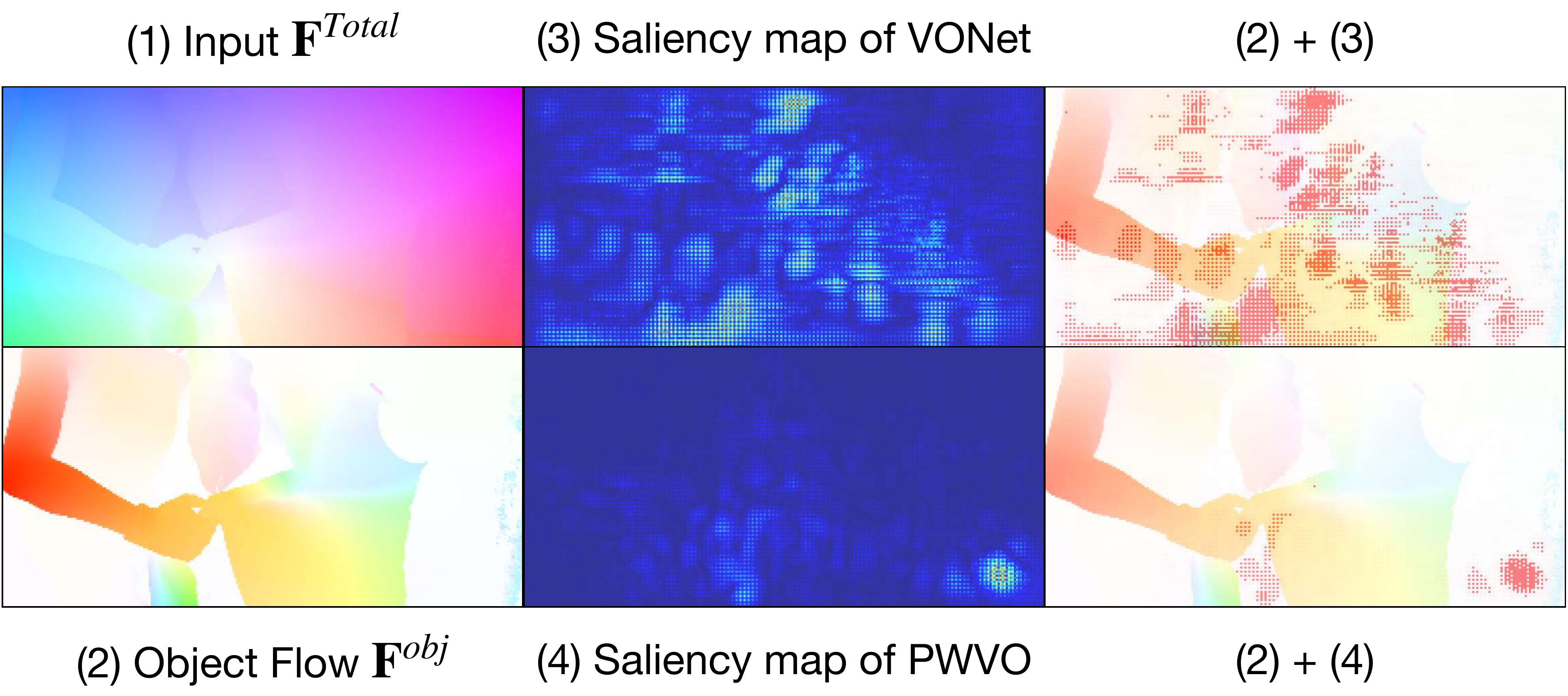}
        \caption{A comparison of VONet and PWVO via the highlighted pixels in their saliency maps.
        }
        \label{fig:saliency}
    \end{minipage}\hfill
    \begin{minipage}[t]{.49\textwidth}
        \centering
        \includegraphics[width=\textwidth]{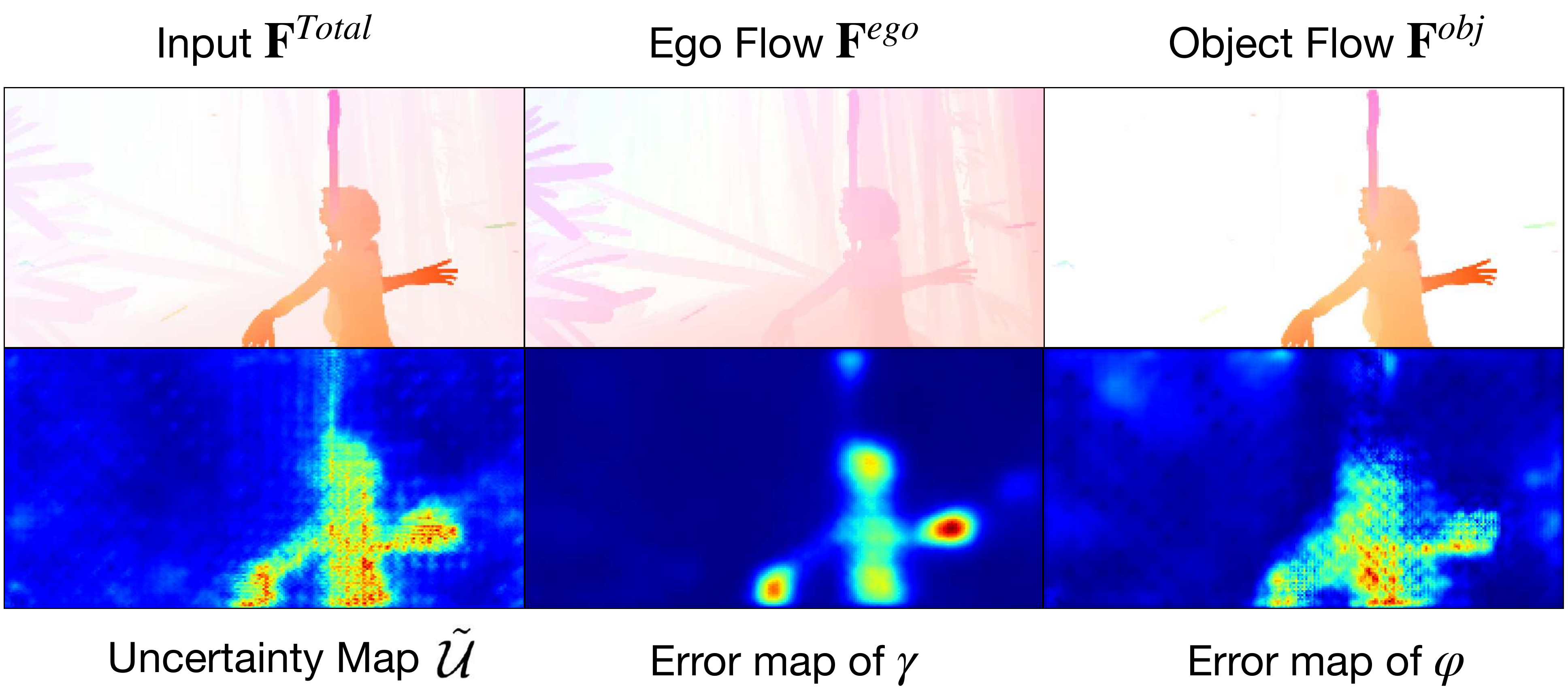}
        \caption{A comparison of the uncertainty map predicted by PWVO v.s. the error maps and $\mathbf{{F}}^{obj}$.}
        \label{fig:uncertainty_map}
    \end{minipage}
\end{figure}


 
\begin{figure}[t]
  \centering
  \includegraphics[width=.95\textwidth]{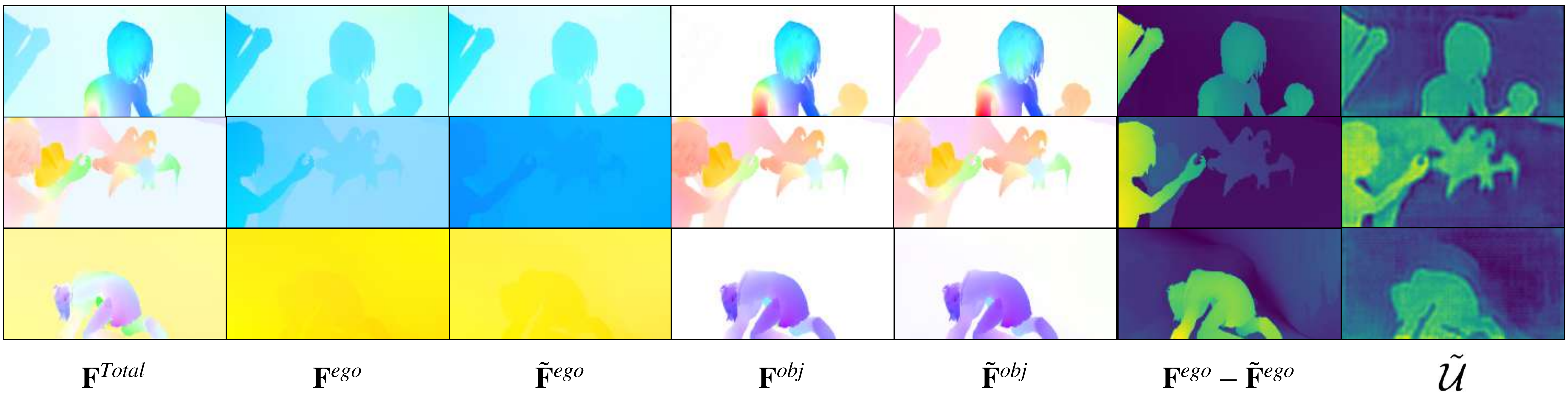}
  \caption{Evaluation of PWVO on the validation set of Sintel. }
  \label{fig:sintel_results}
\end{figure}

\subsection{Qualitative Results}
In this section, we examine the qualitative results for validating the designs adopted by PWVO.

\subsubsection{Examination of the Ability for Dealing with Noises through Saliency Map}
\label{subsub:saliency_map}

Fig.~\ref{fig:saliency} compares PWVO and the baseline VONet from the perspective of their saliency maps, which highlight the pixels that contribute to the predictions of $(\tilde{\gamma}_{i}, \tilde{\varphi}_{i})$.  The first column shows an input $\mathbf{{F}}^{total}$ and its corresponding $\mathbf{{F}}^{obj}$ from Sintel, the second column depicts the saliency maps of VONet and PWVO using their integrated gradients~\citep{simonyan2014deep}, and the third column overlaps the saliency maps with $\mathbf{{F}}^{obj}$.   It can be observed that the highlighted pixels of VONet's saliency map are widely scattered, and cover both the object regions and the background.  In contrast, the highlighted pixels of PWVO's saliency map only fall on the background. This observation thus confirms our hypothesis that PWVO is capable of effectively suppressing the influence of noises when performing VO tasks.

\subsubsection{Examination of the Uncertainty Map Estimated by PWVO}
\label{subsubsec::uncertainty_map}

In order to examine if the uncertainty maps predicted by PWVO can correctly capture moving objects from their background, we further showcase an example selected from Sintel and visualize its input $\mathbf{{F}}^{total}$, $\mathbf{{F}}^{ego}$, and $\mathbf{{F}}^{obj}$ in the first row of Fig.~\ref{fig:uncertainty_map},  as well as the uncertainty map $\tilde{\mathcal{U}}$ and the error maps of $(\tilde{\gamma}_{i}, \tilde{\varphi}_{i})$ predicted by PWVO in the second row of Fig.~\ref{fig:uncertainty_map}.  It can be observed that $\tilde{\mathcal{U}}$ is highly correlated with the error maps and $\mathbf{{F}}^{obj}$, implying that $\tilde{\mathcal{U}}$ is indeed able to capture the noises in the input observations. Since $\tilde{\mathcal{U}}^{\mathcal{R}}$ and $\tilde{\mathcal{U}}^{\mathcal{T}}$ are similar in this case, we only depict one of them in Fig.~\ref{fig:uncertainty_map}.




\subsubsection{Evaluation on the Sintel Validation Set}
\label{subsubsec::evaluation_Sintel}

Fig.~\ref{fig:sintel_results} presents several examples for demonstrating the qualitative evaluation results of PWVO on the Sintel validation set.  It can be observed that the predicted  $\Hat{\mathbf{{F}}}^{ego}$ and  $\Hat{\mathbf{{F}}}^{obj}$ align closely with the ground truth labels, and the estimated uncertainty maps are highly correlated with the error maps of $\Hat{\mathbf{{F}}}^{ego}$. This evidence therefore validates the fact that PWVO trained on the dataset generated by the proposed data generation workflow can deliver favorable results on the validation set of Sintel as well. Please note that visualizations of more examples are provided in the supplementary material.



\section{Limitations and Future Directions}
\label{sec:limitation_and_future_direction}

Abeit effective, PWVO still has limitations in certain scenarios.  For example, in the case that moving objects cover most of the regions in the input observations, PWVO might be misled and treats them as $\Hat{\mathbf{{F}}}^{ego}$. This is due to the lack of sufficient information to derive the motion of the camera, and might also lead to negative impacts on other VO techniques.  Moreover, since PWVO takes $\mathbf{F}^{total}_{i}$, $(\mathbf{D}_{i}, \mathbf{D}_{i+1})$, $\mathbf{x}$, and $\mathbf{K}_{i}$ as its inputs, its performance may degrade if these inputs are not accurate enough. In the future, we plan to further extend PWVO to incorporate random occlusion masks or noises into the inputs, and introduce imperfect input observations to reflect more practical scenarios.






\section{Conclusion}
\label{sec:conclusion}

In this paper, we proposed the concept of utilizing pixel-wise predictions in VO. To achieve this objective, we developed a PWVO framework, which integrates pixel-wise predictions based on the estimated uncertainty maps to derive the final $(\tilde{\gamma}_{i}, \tilde{\varphi}_{i})$.  
In order to provide comprehensive data for training PWVO, we designed a fully configurable data generation workflow for generating synthetic training data. In our experiments, we presented results evaluated on the validation set of Sintel. The results demonstrated that PWVO can outperform the baselines in terms of \textit{$R_{err}$}, \textit{$T_{err}$}, and EPE. In addition, our analyses validated the effectiveness of the components adopted by PWVO, and showed that the designs in PWVO can indeed capture the noises, and suppress the influence from them.





\clearpage
\acknowledgments{If a paper is accepted, the final camera-ready version will (and probably should) include acknowledgments. All acknowledgments go at the end of the paper, including thanks to reviewers who gave useful comments, to colleagues who contributed to the ideas, and to funding agencies and corporate sponsors that provided financial support.}


\bibliography{PWVO}  

\newcommand{\IJCV}{Int. J. Computer Vision (IJCV)}\newcommand{\CVPR}{Proc. IEEE
  Conf. on Computer Vision and Pattern Recognition
  (CVPR)}\newcommand{\CVPRW}{Proc. IEEE Conf. on Computer Vision and Pattern
  Recognition Workshop (CVPRW)}\newcommand{\ICCV}{Proc. IEEE Int. Conf. on
  Computer Vision (ICCV)}\newcommand{\ICCVW}{Proc. IEEE Int. Conf. on Computer
  Vision Workshop (ICCVW)}\newcommand{\ECCV}{Proc. European Conf. on Computer
  Vision (ECCV)}\newcommand{\ECCVW}{Proc. European Conf. on Computer Vision
  (ECCVW)}\newcommand{\IROS}{Proc. IEEE Int. Conf. on Intelligent Robots and
  Systems (IROS)}\newcommand{\CoRL}{Proc. Conf. on Robot Learning
  (CoRL)}\newcommand{\ICRA}{Proc. IEEE Int. Conf. on Robotics and Automation
  (ICRA)}\newcommand{\AAAI}{Proc. AAAI Conf. on Artificial Intelligence
  (AAAI)}\newcommand{\IJCAI}{Proc. Int. Joint Conf. on Artificial Intelligence
  (IJCAI)}\newcommand{\PAMI}{IEEE Trans. Pattern Analysis and Machine
  Intelligence (TPAMI)}\newcommand{\NIPS}{Proc. Conf. on Neural Information
  Processing Systems (NeurIPS)}\newcommand{\ICML}{Proc. Int. Conf. on Machine
  Learning (ICML)}\newcommand{\ICLR}{Proc. Int. Conf. on Learning
  Representations (ICLR)}\newcommand{\ICLRW}{Proc. Int. Conf. on Learning
  Representations Workshop (ICLRW)}\newcommand{\ICASSP}{Proc. IEEE Int. Conf.
  on Acoustics, Speech, & Signal Processing (ICASSP)}\newcommand{\BMVC}{Proc.
  British Machine Vision Conf. (BMVC)}\newcommand{\ACCV}{Proc. Asian Conf. on
  Computer Vision (ACCV)}\newcommand{\WACV}{Proc. IEEE Winter Conf. on
  Applications of Computer Vision (WACV)}
\begin{thebibliography}{56}
\providecommand{\natexlab}[1]{#1}
\providecommand{\url}[1]{\texttt{#1}}
\expandafter\ifx\csname urlstyle\endcsname\relax
  \providecommand{\doi}[1]{doi: #1}\else
  \providecommand{\doi}{doi: \begingroup \urlstyle{rm}\Url}\fi

\bibitem[Konda and Memisevic(2015)]{konda2015learning}
K.~Konda and R.~Memisevic.
\newblock Learning visual odometry with a convolutional network.
\newblock In \emph{{VISAPP} International Conference on Computer Vision Theory
  and Applications}, 2015.

\bibitem[Agrawal et~al.(2015)Agrawal, Carreira, and Malik]{agrawal2015learning}
P.~Agrawal, J.~Carreira, and J.~Malik.
\newblock Learning to see by moving.
\newblock In \emph{\ICCV}, 2015.

\bibitem[Jayaraman and Grauman(2015)]{jayaraman2015learning}
D.~Jayaraman and K.~Grauman.
\newblock Learning image representations tied to ego-motion.
\newblock In \emph{\ICCV}, 2015.

\bibitem[Wang et~al.(2017)Wang, Clark, Wen, and Trigoni]{deepvo}
S.~Wang, R.~Clark, H.~Wen, and N.~Trigoni.
\newblock Deepvo: Towards end-to-end visual odometry with deep recurrent
  convolutional neural networks.
\newblock In \emph{\ICRA}, 2017.

\bibitem[Walch et~al.(2017)Walch, Hazirbas, Leal-Taixé, Sattler, Hilsenbeck,
  and Cremers]{walch2017image}
F.~Walch, C.~Hazirbas, L.~Leal-Taixé, T.~Sattler, S.~Hilsenbeck, and
  D.~Cremers.
\newblock Image-based localization using lstms for structured feature
  correlation.
\newblock In \emph{\ICCV}, 2017.

\bibitem[Xiang et~al.(2018)Xiang, Schmidt, Narayanan, and Fox]{posecnn}
Y.~Xiang, T.~Schmidt, V.~Narayanan, and D.~Fox.
\newblock Posecnn: A convolutional neural network for 6d object pose estimation
  in cluttered scenes.
\newblock In \emph{Proc. Robotics: Science and Systems (RSS)}, 2018.

\bibitem[Balntas et~al.(2018)Balntas, Li, and Prisacariu]{relocnet}
V.~Balntas, S.~Li, and V.~Prisacariu.
\newblock Relocnet: Continuous metric learning relocalisation using neural
  nets.
\newblock In \emph{\ECCV}, 2018.

\bibitem[Laskar et~al.(2017)Laskar, Melekhov, Kalia, and
  Kannala]{laskar2017camera}
Z.~Laskar, I.~Melekhov, S.~Kalia, and J.~Kannala.
\newblock Camera relocalization by computing pairwise relative poses using
  convolutional neural network.
\newblock In \emph{\ICCVW}, 2017.

\bibitem[Melekhov et~al.(2017)Melekhov, Ylioinas, Kannala, and
  Rahtu]{melekhov2017relative}
I.~Melekhov, J.~Ylioinas, J.~Kannala, and E.~Rahtu.
\newblock Relative camera pose estimation using convolutional neural networks.
\newblock In \emph{International Conference on Advanced Concepts for
  Intelligent Vision Systems}, 2017.

\bibitem[Kendall et~al.(2015)Kendall, Grimes, and Cipolla]{posenet}
A.~Kendall, M.~Grimes, and R.~Cipolla.
\newblock Posenet: A convolutional network for real-time 6-dof camera
  relocalization.
\newblock In \emph{\ICCV}, pages 2938--2946, 10 2015.

\bibitem[Kendall and Cipolla(2016)]{posenetwuncertainty}
A.~Kendall and R.~Cipolla.
\newblock Modelling uncertainty in deep learning for camera relocalization.
\newblock In \emph{\ICRA}, 2016.

\bibitem[Cai et~al.(2018)Cai, Shen, and Reid]{cai2018hybrid}
M.~Cai, C.~Shen, and I.~Reid.
\newblock A hybrid probabilistic model for camera relocalization.
\newblock In \emph{\BMVC}, 2018.

\bibitem[Kendall and Cipolla(2017)]{kendall2017geometric}
A.~Kendall and R.~Cipolla.
\newblock Geometric loss functions for camera pose regression with deep
  learning.
\newblock In \emph{\CVPR}, pages 6555--6564, 2017.

\bibitem[Sattler et~al.(2019)Sattler, Zhou, Pollefeys, and
  Leal-Taixé]{sattler2019understanding}
T.~Sattler, Q.~Zhou, M.~Pollefeys, and L.~Leal-Taixé.
\newblock Understanding the limitations of cnn-based absolute camera pose
  regression.
\newblock In \emph{\CVPR}, 2019.

\bibitem[Brahmbhatt et~al.(2018)Brahmbhatt, Gu, Kim, Hays, and
  Kautz]{geomapnet}
S.~Brahmbhatt, J.~Gu, K.~Kim, J.~Hays, and J.~Kautz.
\newblock Geometry-aware learning of maps for camera localization.
\newblock In \emph{\CVPR}, 2018.

\bibitem[Melekhov et~al.(2017)Melekhov, Ylioinas, Kannala, and
  Rahtu]{melekhov2017image}
I.~Melekhov, J.~Ylioinas, J.~Kannala, and E.~Rahtu.
\newblock Image-based localization using hourglass networks.
\newblock In \emph{\ICCVW}, 2017.

\bibitem[Naseer and Burgard(2017)]{naseer2017deep}
T.~Naseer and W.~Burgard.
\newblock Deep regression for monocular camera-based 6-dof global localization
  in outdoor environments.
\newblock In \emph{\IROS}, 2017.

\bibitem[Radwan et~al.(2018)Radwan, Valada, and Burgard]{vlocnet++}
N.~Radwan, A.~Valada, and W.~Burgard.
\newblock Vlocnet++: Deep multitask learning for semantic visual localization
  and odometry.
\newblock In \emph{{IEEE} Robotics Autom. Lett.}, 2018.

\bibitem[Valada et~al.(2017)Valada, Radwan, and Burgard]{valada2017deep}
A.~Valada, N.~Radwan, and W.~Burgard.
\newblock Deep auxiliary learning for visual localization and odometry.
\newblock In \emph{\ICRA}, 2017.

\bibitem[Wu et~al.(2017)Wu, Ma, and Hu]{wu2017delving}
J.~Wu, L.~Ma, and X.~Hu.
\newblock Delving deeper into convolutional neural networks for camera
  relocalization.
\newblock In \emph{\ICRA}, 2017.

\bibitem[Xue et~al.(2018)Xue, Wang, Wang, Dong, Wang, and Zha]{xue2018guided}
F.~Xue, Q.~Wang, X.~Wang, W.~Dong, J.~Wang, and H.~Zha.
\newblock Guided feature selection for deep visual odometry.
\newblock In \emph{\ACCV}, 2018.

\bibitem[Yang et~al.(2020)Yang, von Stumberg, Wang, and Cremers]{d3vo}
N.~Yang, L.~von Stumberg, R.~Wang, and D.~Cremers.
\newblock D3vo: Deep depth, deep pose and deep uncertainty for monocular visual
  odometry.
\newblock In \emph{\CVPR}, 2020.

\bibitem[Costante et~al.(2016)Costante, Mancini, Valigi, and
  Ciarfuglia]{costante2016exploring}
G.~Costante, M.~Mancini, P.~Valigi, and T.~A. Ciarfuglia.
\newblock Exploring representation learning with cnns for frame-to-frame
  ego-motion estimation.
\newblock In \emph{IEEE Robotics Autom. Lett.}, 2016.

\bibitem[Costante and Ciarfuglia(2018)]{lsvo}
G.~Costante and T.~A. Ciarfuglia.
\newblock Ls-vo: Learning dense optical subspace for robust visual odometry
  estimation.
\newblock In \emph{IEEE Robotics Autom. Lett.}, 2018.

\bibitem[Muller and Savakis(2017)]{flowdometry}
P.~Muller and A.~Savakis.
\newblock Flowdometry: An optical flow and deep learning based approach to
  visual odometry.
\newblock In \emph{\WACV}, 2017.

\bibitem[Wang et~al.(2019)Wang, Chen, Lu, Zhao, Trigoni, and
  Markham]{wang2019atloc}
B.~Wang, C.~Chen, C.~X. Lu, P.~Zhao, N.~Trigoni, and A.~Markham.
\newblock Atloc: Attention guided camera localization.
\newblock \emph{arXiv preprint arXiv:1909.03557}, 2019.

\bibitem[Damirchi et~al.(2020)Damirchi, Khorrambakht, and
  Taghirad]{hamed2020attn}
H.~Damirchi, R.~Khorrambakht, and H.~D. Taghirad.
\newblock Exploring self-attention for visual odometry.
\newblock \emph{arXiv}, abs/2011.08634, 2020.

\bibitem[Parisotto et~al.(2018)Parisotto, Chaplot, Zhang, and
  Salakhutdinov]{ParisottoCZS18}
E.~Parisotto, D.~S. Chaplot, J.~Zhang, and R.~Salakhutdinov.
\newblock Global pose estimation with an attention-based recurrent network.
\newblock In \emph{\CVPRW}, pages 237--246, 2018.

\bibitem[Chen et~al.(2019)Chen, Rosa, Miao, Lu, Wu, Markham, and
  Trigoni]{ChenRMLWMT19}
C.~Chen, S.~Rosa, Y.~Miao, C.~X. Lu, W.~Wu, A.~Markham, and N.~Trigoni.
\newblock Selective sensor fusion for neural visual-inertial odometry.
\newblock In \emph{\CVPR}, pages 10542--10551, 2019.

\bibitem[Ciarfuglia et~al.(2014)Ciarfuglia, Costante, Valigi, and
  Ricci]{CiarfugliaCVR14}
T.~A. Ciarfuglia, G.~Costante, P.~Valigi, and E.~Ricci.
\newblock Evaluation of non-geometric methods for visual odometry.
\newblock \emph{Robotics Auton. Syst.}, 62\penalty0 (12):\penalty0 1717--1730,
  2014.

\bibitem[Zhang et~al.(2009)Zhang, Liu, K{\"{u}}hnlenz, and Buss]{ZhangLKB09}
T.~Zhang, X.~Liu, K.~K{\"{u}}hnlenz, and M.~Buss.
\newblock Visual odometry for the autonomous city explorer.
\newblock In \emph{\IROS}, pages 3513--3518, 2009.

\bibitem[Kuo et~al.(2020)Kuo, Liu, Lin, Luo, Chen, and Lee]{DAVO}
X.~Kuo, C.~Liu, K.~Lin, E.~Luo, Y.~Chen, and C.~Lee.
\newblock Dynamic attention-based visual odometry.
\newblock In \emph{\IROS}, pages 5753--5760, 2020.

\bibitem[Kaneko et~al.(2018)Kaneko, Iwami, Ogawa, Yamasaki, and
  Aizawa]{KanekoIOYA18}
M.~Kaneko, K.~Iwami, T.~Ogawa, T.~Yamasaki, and K.~Aizawa.
\newblock Mask-slam: Robust feature-based monocular {SLAM} by masking using
  semantic segmentation.
\newblock In \emph{\CVPRW}, pages 258--266, 2018.

\bibitem[Besc{\'{o}}s et~al.(2018)Besc{\'{o}}s, F{\'{a}}cil, Civera, and
  Neira]{BescosFCN18}
B.~Besc{\'{o}}s, J.~M. F{\'{a}}cil, J.~Civera, and J.~Neira.
\newblock Dynaslam: Tracking, mapping, and inpainting in dynamic scenes.
\newblock \emph{{IEEE} Robotics Autom. Lett.}, 3\penalty0 (4):\penalty0
  4076--4083, 2018.

\bibitem[Sun et~al.(2019)Sun, Sun, Liu, and Yeung]{moav}
T.~Sun, Y.~Sun, M.~Liu, and D.~Yeung.
\newblock Movable-object-aware visual {SLAM} via weakly supervised semantic
  segmentation.
\newblock \emph{CoRR}, abs/1906.03629, 2019.
\newblock URL \url{http://arxiv.org/abs/1906.03629}.

\bibitem[Chen et~al.(2019)Chen, Rosa, Miao, Lu, Wu, Markham, and
  Trigoni]{chen2019selective}
C.~Chen, S.~Rosa, Y.~Miao, C.~X. Lu, W.~Wu, A.~Markham, and N.~Trigoni.
\newblock Selective sensor fusion for neural visual-inertial odometry.
\newblock In \emph{\CVPR}, 2019.

\bibitem[Gao et~al.(2020)Gao, Yu, Shen, Wang, and Yang]{feng2020atten}
F.~Gao, J.~Yu, H.~Shen, Y.~Wang, and H.~Yang.
\newblock Attentional separation-and-aggregation network for self-supervised
  depth-pose learning in dynamic scenes.
\newblock \emph{CoRR}, abs/2011.09369, 2020.

\bibitem[Li et~al.(2022)Li, Wang, Ye, Gong, and Xiang]{bin2022cross}
B.~Li, S.~Wang, H.~Ye, X.~Gong, and Z.~Xiang.
\newblock Cross-modal knowledge distillation for depth privileged monocular
  visual odometry.
\newblock \emph{IEEE Robotics and Automation Letters}, 7\penalty0 (3):\penalty0
  6171--6178, 2022.
\newblock \doi{10.1109/LRA.2022.3166457}.

\bibitem[Lee et~al.(2021)Lee, Rameau, Pan, and Kweon]{lee2021atten}
S.~Lee, F.~Rameau, F.~Pan, and I.~S. Kweon.
\newblock Attentive and contrastive learning for joint depth and motion field
  estimation.
\newblock \emph{CoRR}, abs/2110.06853, 2021.

\bibitem[Kendall and Gal(2017)]{kendall2017what}
A.~Kendall and Y.~Gal.
\newblock What uncertainties do we need in bayesian deep learning for computer
  vision?
\newblock In \emph{\NIPS}, pages 5574--5584, 2017.

\bibitem[Klodt and Vedaldi(2018)]{Klodt2018SupervisingTN}
M.~Klodt and A.~Vedaldi.
\newblock Supervising the new with the old: Learning sfm from sfm.
\newblock In \emph{\ECCV}, 2018.

\bibitem[Strasdat et~al.(2010)Strasdat, Montiel, and Davison]{5509636}
H.~Strasdat, J.~M.~M. Montiel, and A.~J. Davison.
\newblock Real-time monocular slam: Why filter?
\newblock In \emph{2010 IEEE International Conference on Robotics and
  Automation}, pages 2657--2664, 2010.
\newblock \doi{10.1109/ROBOT.2010.5509636}.

\bibitem[Engel et~al.(2013)Engel, Sturm, and Cremers]{6751290}
J.~Engel, J.~Sturm, and D.~Cremers.
\newblock Semi-dense visual odometry for a monocular camera.
\newblock In \emph{2013 IEEE International Conference on Computer Vision},
  pages 1449--1456, 2013.
\newblock \doi{10.1109/ICCV.2013.183}.

\bibitem[Dai et~al.(2021)Dai, Meng, and Jin]{DAI2021107459}
X.-Y. Dai, Q.-H. Meng, and S.~Jin.
\newblock Uncertainty-driven active view planning in feature-based monocular
  vslam.
\newblock \emph{Applied Soft Computing}, 108:\penalty0 107459, 2021.
\newblock ISSN 1568-4946.
\newblock \doi{https://doi.org/10.1016/j.asoc.2021.107459}.
\newblock URL
  \url{https://www.sciencedirect.com/science/article/pii/S1568494621003823}.

\bibitem[Costante and Mancini(2020)]{CostanteM20}
G.~Costante and M.~Mancini.
\newblock Uncertainty estimation for data-driven visual odometry.
\newblock \emph{{IEEE} Trans. Robotics}, 36\penalty0 (6):\penalty0 1738--1757,
  2020.

\bibitem[Mur-Artal and Tardós(2017)]{orbslam2}
R.~Mur-Artal and J.~D. Tardós.
\newblock Orb-slam2: an open-source slam system for monocular, stereo and rgb-d
  cameras.
\newblock In \emph{{IEEE} Trans. Robotics}, 2017.

\bibitem[Klein and Murray(2007)]{klein2007parallel}
G.~Klein and D.~W. Murray.
\newblock Parallel tracking and mapping for small ar workspaces.
\newblock In \emph{{IEEE/ACM} International Symposium on Mixed and Augmented
  Reality (ISMAR)}, 2007.

\bibitem[Geiger et~al.(2011)Geiger, Ziegler, and Stiller]{viso2}
A.~Geiger, J.~Ziegler, and C.~Stiller.
\newblock Stereoscan: Dense 3d reconstruction in real-time.
\newblock In \emph{{IEEE} Intelligent Vehicles Symposium (IV)}, 2011.

\bibitem[Engel et~al.(2017)Engel, Koltun, and Cremers]{dso}
J.~Engel, V.~Koltun, and D.~Cremers.
\newblock Direct sparse odometry.
\newblock \emph{\PAMI}, 2017.

\bibitem[Engel et~al.(2014)Engel, Schöps, and Cremers]{lsdslam}
J.~Engel, T.~Schöps, and D.~Cremers.
\newblock Lsd-slam: Large-scale direct monocular.
\newblock In \emph{\ECCV}, 2014.

\bibitem[Newcombe et~al.(2011)Newcombe, Lovegrove, and Davison]{dtam}
R.~A. Newcombe, S.~Lovegrove, and A.~J. Davison.
\newblock Dtam: Dense tracking and mapping in real-time.
\newblock In \emph{\ICCV}, 2011.

\bibitem[Liu et~al.(2018)Liu, Lehman, Molino, Such, Frank, Sergeev, and
  Yosinski]{coordconv}
R.~Liu, J.~Lehman, P.~Molino, F.~P. Such, E.~Frank, A.~Sergeev, and
  J.~Yosinski.
\newblock An intriguing failing of convolutional neural networks and the
  coordconv solution.
\newblock In \emph{\NIPS}, pages 9628--9639, 2018.

\bibitem[Fischler and Bolles(1981)]{pnp}
M.~A. Fischler and R.~C. Bolles.
\newblock Random sample consensus: A paradigm for model fitting with
  applications to image analysis and automated cartography.
\newblock \emph{Commun. ACM}, 24\penalty0 (6):\penalty0 381–395, jun 1981.
\newblock ISSN 0001-0782.
\newblock \doi{10.1145/358669.358692}.
\newblock URL \url{https://doi.org/10.1145/358669.358692}.

\bibitem[Butler et~al.(2012)Butler, Wulff, Stanley, and Black]{sintel}
D.~J. Butler, J.~Wulff, G.~B. Stanley, and M.~J. Black.
\newblock A naturalistic open source movie for optical flow evaluation.
\newblock In \emph{\ECCV}, pages 611--625, 2012.

\bibitem[Wang et~al.(2020)Wang, Hu, and Scherer]{tartan}
W.~Wang, Y.~Hu, and S.~A. Scherer.
\newblock Tartanvo: A generalizable learning-based vo.
\newblock In \emph{\CoRL}, 2020.

\bibitem[Simonyan et~al.(2014)Simonyan, Vedaldi, and
  Zisserman]{simonyan2014deep}
K.~Simonyan, A.~Vedaldi, and A.~Zisserman.
\newblock Deep inside convolutional networks: Visualising image classification
  models and saliency maps.
\newblock In \emph{\ICLRW}, 2014.

\end{thebibliography}

\end{document}